\documentclass[]{interact}
\usepackage{epstopdf}% To incorporate .eps illustrations using PDFLaTeX, etc.
\usepackage[caption=false]{subfig}% Support for small, `sub' figures and tables

\usepackage[numbers,sort&compress]{natbib}% Citation support using natbib.sty
\bibpunct[, ]{[}{]}{,}{n}{,}{,}% Citation support using natbib.sty
\makeatletter% @ becomes a letter
\def\NAT@def@citea{\def\@citea{\NAT@separator}}% Suppress spaces between citations using natbib.sty
\makeatother% @ becomes a symbol again

\theoremstyle{plain}% Theorem-like structures provided by amsthm.sty
\newtheorem{theorem}{Theorem}[section]

\newtheorem{proposition}[theorem]{Proposition}

\theoremstyle{definition}
\newtheorem{definition}[theorem]{Definition}

\theoremstyle{remark}

\usepackage{algorithm,algorithmic}
\usepackage{epstopdf}
\graphicspath{{pics/}{fig/}}

\begin{document}
%\articletype{ARTICLE TEMPLATE}
\title{A fast method for simultaneous reconstruction and segmentation in X-ray CT application}

\author{
\name{Yiqiu Dong\textsuperscript{a,b} Chunlin Wu\textsuperscript{c} and Shi Yan\textsuperscript{a,c}\thanks{CONTACT: Shi Yan. Email: shi.yan@ensicaen.fr . Shi Yan is now with Normandie Univ. UniCaen, ENSICAEN, CNRS, GREYC, France.}}
\affil{
\textsuperscript{a} Department of Applied Mathematics and Computer Science, Technical University of Denmark, 2800 Kgs. Lyngby, Denmark.;
\textsuperscript{b} College of Mathematics and Statistics, Shenzhen University, Shenzhen, Guangdong, P.R.China.
\textsuperscript{c} School of Mathematical Science, Nankai University, 300071 Tianjin, China.;
}
}

\maketitle

\begin{abstract}
In this paper, we propose a fast method for simultaneous reconstruction and segmentation (SRS) in X-ray computed tomography (CT). 
Our work is based on the SRS model where Bayes' rule and the maximum a posteriori (MAP) are used on hidden Markov measure field model (HMMFM). 
The original method leads to a logarithmic-summation (log-sum) term, which is non-separable to the classification index. The minimization problem in the model was solved by using constrained gradient descend method, Frank-Wolfe algorithm, which is very time-consuming especially when dealing with large-scale CT problems.
The starting point of this paper is the commutativity of log-sum operations, where the log-sum problem could be transformed into a sum-log problem by introducing an auxiliary variable. The corresponding sum-log problem for the SRS model is separable. After applying alternating minimization method, this problem turns into several easy-to-solve convex sub-problems. 
In the paper, we also study an improved model by adding Tikhonov regularization, and give some convergence results. 
Experimental results demonstrate that the proposed algorithms could produce comparable results with the original SRS method with much less CPU time. 
\end{abstract}


\section{Introduction}
X-ray computed tomography (CT) is an important application of inverse problems, which reconstructs the attenuation coefficients of an object from the damping of X-rays. Since different materials have different attenuation coefficients, by X-ray CT technique we are able to see interior of the object. As X-ray CT is widely applied in a lot of fields, many reconstruction methods have been proposed and applied in the industry, such as the filtered back projection algorithm \cite{kuchment2014radon}, and algebraic reconstruction techniques \cite{kak2001principles,hansen2012air}.

After reconstruction, very often we would apply image segmentation technique to distinguish some interested regions or objects. A main drawback of separating reconstruction and segmentation is the error propagation, i.e., the errors in the reconstruction will continue to affect the segmentation. To overcome this drawback, several simultaneous reconstruction and segmentation (SRS) methods were proposed in recent years. As far as we know, the first SRS method in CT was proposed in \cite{ramlau2007mumford} based on Mumford-Shah level-set approach \cite{mumford1989optimal}.
In \cite{van2008simultaneous}, another SRS method according to hidden Markov measure field model (HMMFM) \cite{marroquin2003hidden} was proposed, where the segmentation result is obtained through a probability map. For CT with shadowed data, an SRS method based on Potts model \cite{potts1952some} was proposed in \cite{10.1007/978-3-319-58771-4_25}. In order to segment the objects that have different patterns during CT reconstruction, in \cite{8417955} the dictionary learning technique was introduced into SRS method.

Following the idea from \cite{marroquin2003hidden}, in \cite{doi:10.1080/17415977.2015.1124428} the means and variances of the segmentation classes were used as prior and a new SRS method was proposed. Numerical results show that this SRS method can significantly improve the accuracy of reconstruction and segmentation. Further, this method was extended by withdrawing the prior information on the variance, see \cite{10.1007/978-3-319-58771-4_21}. In \cite{doi:10.1080/17415977.2015.1124428}, by using Bayes' rule and the maximum a posteriori (MAP) estimate with proper prior information on the means and variances of the classes, the authors proposed the following variational model:
\begin{equation}\label{model_SRS_ori}
\underset{\bm{x}, \bm\delta}{\min}\  E_0(\bm{x},\bm{\delta}) = \lambda_{n}||A\bm{x}-\bm{b}||^2_2 + \lambda_{c}\sum_{k=1}^K R(\bm\delta_k) -\sum_{j=1}^{N}\ln{\left[\sum_{k=1}^K \frac{\delta_{jk}}{\sqrt{2\pi} \sigma_k} \exp(-\frac{(x_j-\mu_k)^2}{2\sigma_k^2})\right]},
\end{equation}
where $A\in \mathbb{R}^{M\times N}$ is the system matrix of CT scanner, $\bm{x}\in \mathbb{R}^N$ is the image of attenuation coefficients, and $\bm{b}\in \mathbb{R}^M$ contains all measurements. Here, we assume that the image consists of $K$ classes, and define $\bm\delta =(\bm\delta_1,\cdots,\bm\delta_K)= \{\delta_{jk}\}\in \mathbb{R}^{N\times K}$ with $\delta_{jk}$ as the probability of pixel $x_j$ belonging to the $k$th class. Then we have the constraint, $\sum_{1}^{K}\delta_{jk} = 1$ and $\delta_{jk}\geq0$ for all $j$ and $k$. Furthermore, $\mu_k$ and $\sigma_k$ are the mean and standard deviation of the $k$th class, respectively. In addition, $\lambda_{n}, \lambda_{c}>0$ are the regularization parameters, and $R(\cdot)$ denotes the regularization according to the prior information. In \cite{doi:10.1080/17415977.2015.1124428}, Tikhonov regularization \cite{tikhonov1943stability} and total variation (TV) regularization \cite{rudin1992nonlinear} have been tested. In the last term of \eqref{model_SRS_ori}, which is called as \textit{log-sum term}, because the logarithmic operator and summation operator are non-separable, the constrained minimization problem in \eqref{model_SRS_ori} is very difficult to solve. When solving $\bm{x}$ in \cite{doi:10.1080/17415977.2015.1124428}, the authors have to use an approximation of the log-sum term.
Moreover, the algorithm is time-consuming, which limits its applications to large-scale CT problems. 

In this paper, we propose a new method to solve the model \eqref{model_SRS_ori}. Inspired by the work in \cite{teboulle2007unified,liu2013weighted}, the log-sum term can be transformed into a \textit{sum-log term} by introducing an auxiliary variable, then the sub-problems become much easier to solve. Numerical results show that our method can provide comparable results as in \cite{doi:10.1080/17415977.2015.1124428} with much less CPU time. 

The rest of the paper is organized as follows. In Section \ref{section_bg}, the method in \cite{doi:10.1080/17415977.2015.1124428} is briefly reviewed. In Section \ref{section_exsolver}, by transforming the log-sum term, we introduce a new method that can solve the minimization problem in the model \eqref{model_SRS_ori} without simplification. Furthermore, an improved model is proposed and some convergence results are given.
In Section \ref{section_exp}, numerical results are presented. Finally, conclusions are drawn in Section \ref{section_con}.

\section{The method in \cite{doi:10.1080/17415977.2015.1124428}}\label{section_bg}

In \cite{doi:10.1080/17415977.2015.1124428}, the authors use alternating minimization method \cite{csiszar1984information} to solve the model \eqref{model_SRS_ori} as follows,

\begin{align}
\bm{x}^{n+1} &= \underset{\bm{x}}{\arg\min}~E_0(\bm{x},\bm\delta^n)\label{ori_alter1},\\
\bm{\delta}^{n+1} &= \underset{\bm{\delta}}{\arg\min}~E_0(\bm{x}^{n+1},\bm\delta)\label{ori_alter2}.
\end{align}

The last term in \eqref{model_SRS_ori} is a log-sum term, which is non-separable to the classification index $k$, thus it is very difficult to minimize the sub-problem according to $\bm{x}$. 
In order to solve the sub-problem \eqref{ori_alter1}, i.e., 
\begin{equation*}\label{model_SRS_ori_x}
\underset{\bm{x}}{\min}\  \lambda_{n}||A\bm{x}-\bm{b}||^2_2 -\sum_{j=1}^{N}\ln{\left[\sum_{k=1}^K \frac{\delta^n_{jk}}{\sqrt{2\pi} \sigma_k} \exp(-\frac{(x_j-\mu_k)^2}{2\sigma_k^2})\right]},
\end{equation*}
the authors applied two-step approximation to simplify the log-sum term.
Denote 
\begin{equation}
p(x_j|\bm\delta_j,\bm\mu,\bm\sigma) = \sum_{k=1}^K \frac{\delta_{jk}}{\sqrt{2\pi} \sigma_k} \exp(-\frac{(x_j-\mu_k)^2}{2\sigma_k^2}).
\label{def_px}
\end{equation}
In the first step, $p(x_j|\bm\delta_j,\bm\mu,\bm\sigma)$ is approximated by a ``flat'' Gaussian distribution
\begin{equation}\label{approximate_1}
\hat{p}(x_j|\bm\delta_j,\bm\mu,\bm\sigma)  = \frac{1}{\sqrt{2\pi} \hat{\sigma}_j}\exp{(- \frac{(x_j-\hat{\mu}_j)^2}{2\hat\sigma^2_j})},
\end{equation}
where 
$$\hat\mu_j = \sum_{k=1}^K \delta_{jk}\mu_k\quad\mbox{and}\quad \hat\sigma^2_j = \sum_{k=1}^K \delta_{jk}(\sigma^2_k + \mu_k^2) - \hat\mu_j^2.$$
After a few iterations on \eqref{ori_alter1} and \eqref{ori_alter2}, as the segmentation $\bm{\delta}$ is good enough, a ``sharp'' Gaussian distribution is used to approximate $p(x_j|\bm\delta_j,\bm\mu,\bm\sigma)$ to further improve the reconstruction result, i.e.,
\begin{equation}\label{approximate_2}
p(x_j|\bm\delta_j,\bm\mu,\bm\sigma) \approx {\tilde{p}(x_j|\bm\delta_j,\bm\mu,\bm\sigma)}  = \frac{1}{\sqrt{2\pi} {\sigma}_{k_j}}\exp{(- \frac{(x_j-{\mu}_{k_j})^2}{2\sigma^2_{k_j}})},
\end{equation}
where $k_j  = \underset{k}{\arg\max}\ \delta_{jk}$.

Clearly, after using a single Gaussian distribution to approximate the sum of several Gaussian distribution in $p(x_j|\bm\delta_j,\bm\mu,\bm\sigma)$, the summation operator of the log-sum in the model \eqref{model_SRS_ori} is eliminated, and the reconstruction $\bm{x}$ is obtained by solving a simple least-squares problem. 

The sub-problem \eqref{ori_alter2} is 
\begin{equation*}\label{model_SRS_ori_delta}
\underset{\bm\delta}{\min}\    \lambda_{c}\sum_{k=1}^K R(\bm\delta_k) -\sum_{j=1}^{N}\ln{\left[\sum_{k=1}^K \frac{\delta_{jk}}{\sqrt{2\pi} \sigma_k} \exp(-\frac{(x_j^n-\mu_k)^2}{2\sigma_k^2})\right]},
\end{equation*}
with the constraint $\sum_{1}^{K}\delta_{jk} = 1$ and $\delta_{jk}\geq0$ for all $j$ and $k$.
Because of the constraint and the non-differentiable TV term, a general constrained gradient descent method such as Frank-Wolfe algorithm \cite{bertsekas1999nonlinear} could be applied here. Due to the singularity of the TV term, it is well known that gradient descent algorithm is very inefficient \cite{Vogel1996Iterative}. 

Obviously, solving the minimization problem in \eqref{model_SRS_ori} is challenging due to the log-sum term. In this paper, we will take advantage of the commutativity of the log-sum operations studied in \cite{teboulle2007unified,liu2013weighted}, and transform the model \eqref{model_SRS_ori} into several easy-to-solve convex sub-problems. Besides, we could provide some convergence results that were missing in \cite{doi:10.1080/17415977.2015.1124428}.

\section{Our method and convergence results}\label{section_exsolver}

In this section, we propose a new method that solves the model \eqref{model_SRS_ori} and provide some convergence results of the new method.

\subsection{Proposed method}
Due to the log-sum term in \eqref{model_SRS_ori}, the sub-problem with respect to $\bm x$ is non-convex and very difficult to solve. Inspired by the transformation of the log-sum operations introduced in \cite{teboulle2007unified,liu2013weighted}, we can transform the log-sum term into a much simpler form. To do so, we would need the following proposition.

\begin{proposition}\label{logsum_lemma}
(commutativity of the log-sum operations \cite{teboulle2007unified,liu2013weighted}). Given $f_k>0, k=1,2\cdots,K$, we have
\begin{equation*}
-\ln \sum_{k=1}^{K} f_k = \min_{\bm\phi\in \mathcal{A}}\{-\sum_{k=1}^K \phi_{k}\ln f_k + \sum_{k=1}^{K}\phi_{k} \ln \phi_{k}\},
\end{equation*}
where $\bm\phi = (\phi_{1},\phi_{2},\cdots,\phi_{K})$, and 
\begin{equation*}\label{def_A}
\mathcal{A} = \{\bm\xi| \sum_{k=1}^{K}\xi_{k}=1, \xi_{k}\in (0,1) \mbox{ for all } k\}.
\end{equation*}
\end{proposition}

According to the log-sum term in \eqref{model_SRS_ori}, we define
\begin{equation}\label{def_f_k}
f_{jk}(x_j,\delta_{jk}) = \frac{\delta_{jk}}{\sqrt{2\pi}\sigma_k} \exp{(-\frac{(x_j-\mu_k)^2}{2\sigma_k^2})}.
\end{equation}

Then, applying Proposition \ref{logsum_lemma} on \eqref{model_SRS_ori}, we transform the minimization problem \eqref{model_SRS_ori} into a new optimization problem with respect to the variables $(\bm x, \bm\delta, \bm\phi)$:
\begin{equation}\label{model_ori_emsolver}
\min_{\bm{x},\bm\delta\in \mathcal{B},\bm\phi\in {\mathcal{B}}}\,  
\left\{
\begin{array}{rcl}
E(\bm{x},\bm\delta,\bm\phi) &=& \lambda_{n}||\bm A\bm x-\bm b||_{2}^2 + \lambda_{c} \displaystyle\sum_{k=1}^{K} R(\bm\delta_k) \\
&&+ \displaystyle\sum_{j=1}^{N}\left[ -\displaystyle\sum_{k=1}^{K} \phi_{jk} \ln f_{jk}(x_j,\delta_{jk}) + \displaystyle\sum_{k=1}^K \phi_{jk} \ln \phi_{jk} \right]
\end{array}
\right\},
\end{equation}
where $\bm\phi\in\mathbb{R}^{N\times K}$ and $\mathcal{B} = \{\bm \xi\in\mathbb{R}^{N\times K}| \sum_{k=1}^K\xi_{jk} = 1 \mbox{ and } \xi_{jk}\in(0,1) \mbox{ for all }j, k\}$. Note that here we let $\delta_{jk}\in (0,1)$ to avoid  $f_{jk}(x_j,\delta_{jk})$ being 0. Furthermore, in our method we use TV regularization on each column of $\bm\delta$, i.e.

\begin{equation*}
R_{TV}(\bm\delta_k) = \sum_{j=1}^N \sqrt{(-\delta_{jk} + \delta_{j'k})^2 + (-\delta_{jk} + \delta_{j''k})^2},
\end{equation*}
where $j', j''$ represent the neighbor pixels {of point $j$} in the horizontal and vertical directions, respectively. At the boundary of the image domain, the value of the nearest pixel is repeated.

In order to solve the minimization problem \eqref{model_ori_emsolver}, we apply alternating minimization method \cite{csiszar1984information}, i.e., we solve the sub-problems with respect to $\bm x$, $\bm\delta$ and $\bm\phi$ alternately. The detailed algorithm is given in Algorithm \ref{alg1}.

\begin{algorithm}[t]
\caption{Algorithm for solving the minimization problem in (\ref{model_ori_emsolver})}
\begin{algorithmic}\label{alg1}
\STATE 1. Set $\lambda_{n}$ and $\lambda_{c}$, and initialize $n=0$, $\bm{x}^0 = 0$, $\bm\delta^0$ and $\bm\phi^0$ with $\delta_{j,k}^0 = \phi_{j,k}^0=\frac{1}{K}$ for all $j$ and $k$. 

\STATE 2. Update $\bm{x}^{n+1}$, by

\begin{equation}\label{x_sub_problem}
\begin{array}{rl}
\bm x^{n+1} =& \underset{\bm{x}}{\arg\min} 
~E (\bm{x},\bm\delta^n,\bm\phi^n)\\
=&\underset{\bm{x}}{\arg\min} \left\{ 
\lambda_{n}||\bm A\bm x-\bm b||^2_2 +  \displaystyle\sum_{j=1}^{N} \sum_{k=1}^{K}\left[  \frac{\phi_{jk}^n}{2\sigma_k^2} {(x_j-\mu_k)^2}\right] \right\}.
\end{array}
\end{equation}
\STATE 3. Update $\bm\delta^{n+1}$, by
\begin{equation}\label{delta_sub_problem}
\begin{array}{rl}
\bm\delta^{n+1} =& \underset{\bm\delta\in \mathcal{B}}{\arg\min}
~E (\bm{x}^{n+1},\bm\delta,\bm\phi^n)\\
=&\underset{\bm\delta\in \mathcal{B}}{\arg\min} \left\{  \lambda_{c} \displaystyle\sum_{k=1}^{K} R_{TV}(\bm\delta_k) + \sum_{j=1}^{N}\sum_{k=1}^{K}(- \phi_{jk}^n\ln\delta_{jk})  \right\}.%
\end{array}
\end{equation}
\STATE 4. Update $\bm\phi^{n+1}$, by
\begin{equation}\label{phi_sub_problem}  
\begin{array}{rl}
\bm\phi^{n+1} = & \underset{\bm\phi\in \mathcal{B}}{\arg\min} 
~E (\bm{x}^{n+1},\bm\delta^{n+1},\bm\phi) \\
=&   \underset{\bm\phi\in \mathcal{B}}{\arg\min} \left\{ 
\displaystyle\sum_{j=1}^{N}\left[ -\sum_{k=1}^{K} \phi_{jk} \ln f_{jk}(x_j^{n+1},\delta_{jk}^{n+1}) + \sum_{k=1}^K \phi_{jk} \ln \phi_{jk} \right]
\right\}.
\end{array}
\end{equation}
\STATE 5. If $\frac{||\bm{x}^{n+1}-\bm{x}^{n}||_2}{||\bm{x}^{n}||_2}<10^{-4}$, then stop. Otherwise, let $n=n+1$, and go to 2.
\end{algorithmic}
\end{algorithm}

In Algorithm \ref{alg1}, the objective function in the sub-problem (\ref{x_sub_problem}) on $\bm x$ is quadratic, so it can be efficiently solved by using CGLS method \cite{bjorck1996numerical}. 
Since the sub-problem (\ref{phi_sub_problem}) is separable, we can solve $\phi_{j,k}$ element-wise. According to the first-order optimality condition together with Lagrange multiplier technique, we can easily obtain the closed-form solution:
\begin{equation}\label{solver_phi}
\phi_{jk}^{n+1} = \frac{f_{jk}(x^n_j,\delta^n_{jk})}{\sum_{l=1}^{K} f_{jl}(x^n_j,\delta^n_{jl})}.
\end{equation}

In the following subsection, we will focus on solving the $\bm\delta$ sub-problem (\ref{delta_sub_problem}). 
\subsubsection{The $\bm\delta$ sub-problem }

The difficulties of solving the minimization problem in (\ref{delta_sub_problem}) are mainly from the non-differentiable term $R_{TV}(\bm\delta_{k})$, highly nonlinear and nonquadratic term $\ln\delta_{jk}$, and the constraint $\bm\delta\in \mathcal{B}$. In order to split them apart, we introduce two auxiliary variables, $\bm\eta, \bm\psi\in \mathbb{R}^{N\times K}$, and let $\bm\delta=\bm\eta =  \bm\psi$, {while each of them take one of the difficulties from $R_{TV}(\bm\delta_{k})$, $\ln\eta_{jk}$, and $\bm\psi\in \mathcal{B}$, respectively.} Now (\ref{delta_sub_problem}) becomes

\begin{equation}
\{\bm\delta^{n+1},\bm\eta^{n+1},\bm\psi^{n+1}\} =
\underset{\bm\psi\in \mathcal{B},\bm\eta = \bm\delta = \bm\psi}{\arg\min} \left\{  \lambda_{c} \sum_{k=1}^{K}  R_{TV}(\bm\delta_k) +
\sum_{k=1}^{K}\sum_{j=1}^{N}(- \phi^n_{jk} \ln\eta_{jk})  \right\}.
\end{equation}

Then, we apply the alternating direction method of multiplier (ADMM) \cite{boyd2011distributed} to solve it. 

By introducing two Lagrangian multipliers $\bm \lambda_{1}, \bm\lambda_{2}\in\mathbb{R}^{N\times K}$ for the linear constraint $\bm\delta=\bm\eta$ and $\bm\eta=\bm\psi$, respectively, we obtain the augmented Lagrangian \cite{Wu2010}: 
\begin{align}
{L}^n(\bm\delta,\bm\eta,\bm\psi,\bm\lambda_1,\bm\lambda_2) =&\ 
\lambda_{c} \sum_{k=1}^{K} R_{TV}(\bm\delta_k) +
\sum_{j=1}^{N} \sum_{k=1}^{K} \left(-\phi^n_{jk} \ln\eta_{jk}\right) \notag \\
&+\frac{\gamma_1}{2} \|\bm\delta - \bm\eta\|^2_F + \langle \bm\lambda_1,\bm\delta - \bm\eta \rangle
+\frac{\gamma_2}{2} ||\bm\eta - \bm\psi||^2_F + \langle \bm\lambda_2, \bm\eta - \bm\psi \rangle, \notag
\end{align}
where $\gamma_{1}$ and $\gamma_{2}$ are positive penalty parameters. According to ADMM, we need solve 
\begin{equation*}
\min\limits_{\bm\delta, \bm\eta, \bm\psi\in \mathbb{R}^{N\times K}}\ \max\limits_{\bm\lambda_1,\bm\lambda_2\in \mathbb{R}^{N\times K}} \ {L}^n(\bm\delta,\bm\eta,\bm\psi,\bm\lambda_1,\bm\lambda_2),
\end{equation*}
in order to obtain $\bm\delta^{n+1}$ in Algorithm \ref{alg1}, and the iterates in ADMM are generated as follows:
\begin{align}
\bm\delta^{m+1} = &\ \underset{\bm\delta}{\arg\min} \left\{\lambda_{c} \sum_{k=1}^{K} R_{TV}(\bm\delta_k) +
\frac{\gamma_1}{2} ||\bm\delta - \bm\eta^m||^2_F + \langle \bm\lambda_1^m, \bm\delta - \bm\eta^m \rangle\right\},\label{subproblem_delta_delta}\\
\bm\eta^{m+1} = & \ \underset{\bm\eta}{\arg\min} \bigg\{ \sum_{j=1}^{N} \sum_{k=1}^{K} \left(-\phi^n_{jk} \ln\eta_{jk}\right) + \frac{\gamma_1}{2} ||\bm\delta^{m+1} - \bm\eta||^2_F + \langle \bm\lambda_1^{m},\bm\delta^{m+1} - \bm\eta \rangle \notag\\ 
& \qquad\qquad +\frac{\gamma_2}{2} ||\bm\eta - \bm\psi^{m}||^2_F + \langle \bm\lambda_2^{m}, \bm\eta - \bm\psi^{m} \rangle\bigg\},\label{subproblem_delta_mu}\\
\bm\psi^{m+1} = &\  \underset{\bm\psi\in\mathcal{B}}{\arg\min}\left\{ \frac{\gamma_2}{2} ||\bm\eta^{m+1} - \bm\psi||^2_F + \langle \bm\lambda^m_2, \bm\eta^{m+1} - \bm\psi \rangle\right\},\label{subproblem_delta_eta}\\
\bm\lambda_1^{m+1} = &\ \bm\lambda_1^{m} + \gamma_1 (\bm\delta^{m+1}-\bm\eta^{m+1}),\notag\\
\bm\lambda_2^{m+1} = &\ \bm\lambda_2^{m} + \gamma_2 (\bm\eta^{m+1}-\bm\psi^{m+1}).\notag
\end{align}

For sub-problem (\ref{subproblem_delta_delta}) we can solve it column-by-column on $\bm\delta$, and each column can be solved by applying split Bregman method \cite{goldstein2009split} with Neumann boundary condition. In sub-problem (\ref{subproblem_delta_mu}), each element $\eta_{jk}$ of $\bm\eta$ can be solved separately. Based on the first-order optimaility condition, we obtain that the minimizer $\eta_{jk}$ should satisfy the following equation

\begin{equation*}\label{solver_delta_mu}
-\frac{\phi_{jk}^n}{\eta_{jk}} +{\gamma_1}(\eta_{jk}-\delta^{m+1}_{jk}) -\lambda^m_{1jk} + {\gamma_2}(\eta_{jk} - \psi^m_{jk}) + \lambda^m_{2jk}=0.
\end{equation*}
Due to the logarithmic operation, $\eta_{jk}$ need be positive, and its solution has closed form: 

\begin{equation}
\eta_{jk}^{m+1} =  \frac{\gamma_1\delta_{jk}^{m+1} + \lambda_{1jk}^m + \gamma_2 \psi_{jk}^m - \lambda_{2jk}^m+\sqrt{\Delta_{jk}^{m}}}{2(\gamma_1+\gamma_2)},
\end{equation}
with
\begin{equation*}
\Delta_{jk}^{m} = (\gamma_1\delta_{jk}^{m+1} + \lambda_{1jk}^m + \gamma_2 \psi_{jk}^m - \lambda_{2jk}^m)^2+4\phi_{jk}^n(\gamma_1+\gamma_2).
\end{equation*}

Sub-problem (\ref{subproblem_delta_eta}) is a least-squares problem with a constraint on a convex set, and one approximate solution could be obtained by
\begin{equation}
\psi^{m+1}_{jk}=\frac{\max\{\gamma_{2}\eta^{m+1}_{jk}+{\lambda^{m}_{2}}_{jk},\epsilon\}}{\sum_{s=1}^{k}(\max\{\gamma_{2}\eta^{m+1}_{js}+{\lambda^{m}_{2}}_{js},\epsilon\})},
\end{equation}
where $\epsilon$ is a pre-defined small positive number.

\subsection{An improved model based on (\ref{model_ori_emsolver})}\label{section_impmodel}

If we look back to the model (\ref{model_ori_emsolver}), we can see that the only regularization term in the model is on $\bm\delta$.
In \cite{doi:10.1080/17415977.2015.1124428}, due to the log-sum term in \eqref{model_SRS_ori}, 
the segmentation information represented in $\bm\delta$ is utilized to regularize the reconstruction, and there were not any smoothing requirement on $\bm x$.  While in the proposed new model \eqref{x_sub_problem}, $\bm{x}$ is updated by using the unregularized $\bm\phi$, which brings a risk: the isolated points on unregularized $\bm\phi$ might lead to isolated points in the reconstruction result $\bm{x}$. This phenomenon is presented in Figure \ref{8classcompareseg} in our numerical results. In order to further improve the reconstruction, we add Tikhonov regularization on $\bm x$, and modify the model (\ref{model_ori_emsolver}) as follows:
\begin{equation}\label{model_regx}
\min_{\bm{x},\bm\delta\in \mathcal{B},\bm\phi\in \mathcal{B}} 
\left\{
\begin{array}{rl}
F(\bm{x},\bm\delta,\bm\phi) = & 
\lambda_{n}||\bm A\bm x-\bm b||^2_{2} + \lambda_{t} \|\nabla \bm x\|^{2}_{2} + 
\lambda_{c} \displaystyle\sum_{k=1}^{K} R_{TV}(\bm\delta_k)  \\ 
& + \displaystyle\sum_{j=1}^{N}\left[ -\displaystyle\sum_{k=1}^{K} \phi_{jk} \ln f_{jk}(x_j,\delta_{jk}) + \displaystyle\sum_{k=1}^K \phi_{jk} \ln \phi_{jk} \right]
\end{array}
\right\},
\end{equation}
where $\nabla\in\mathbb{R}^{2N\times N}$ denotes the discrete gradient operator using the forward finite difference scheme with Neumann boundary condition, and $\lambda_t>0$ is a regularization parameter. 

\begin{algorithm}[t]
\caption{Algorithm for solving the minimization problem in (\ref{model_regx})}
\begin{algorithmic}\label{alg2}
\STATE 1. Set $\lambda_{n}$ and $\lambda_{c}$, and initialize $\bm{x}^0 = 0$, $\bm\delta^0$ and $\bm\phi^0$ with all $\delta_{j,k}^0 = \phi_{j,k}^0=\frac{1}{K}$. 

\STATE 2. Update $\bm{x}^{n+1}$, by
\begin{equation}\label{x_sub_problem_reg}
\begin{array}{rl}
\bm x^{n+1} =& \underset{\bm{x}}{\arg\min} 
~F (\bm{x},\bm\delta^n,\bm\phi^n)\\
=&\underset{\bm{x}}{\arg\min} \left\{ 
\lambda_{n}||\bm A\bm x-\bm b||^2_2 +  \displaystyle\sum_{j=1}^{N} \sum_{k=1}^{K}\left[  \frac{\phi_{jk}^n}{2\sigma_k^2} {(x_j-\mu_k)^2}\right] + \lambda_{t} \|\nabla \bm x\|^{2}_{2} \right\}.
\end{array}
\end{equation}
\STATE 3. Update $\bm\delta^{n+1}$, by
\begin{equation}\label{delta_sub_problem_reg}
\begin{array}{rl}
\bm\delta^{n+1} =& \underset{\bm\delta\in \mathcal{B}}{\arg\min}
~F (\bm{x}^{n+1},\bm\delta,\bm\phi^n)\\
=&\underset{\bm\delta\in \mathcal{B}}{\arg\min} \left\{  \lambda_{c} \displaystyle\sum_{k=1}^{K} R_{TV}(\bm\delta_k) + \sum_{j=1}^{N}\sum_{k=1}^{K}(- \phi_{jk}^n\ln\delta_{jk})  \right\}.
\end{array}
\end{equation}
\STATE 4. Update $\bm\phi^{n+1}$, by
\begin{equation}\label{phi_sub_problem_reg} 
\begin{array}{rl}
\bm\phi^{n+1} = & \underset{\bm\phi\in \mathcal{B}}{\arg\min} 
~F (\bm{x}^{n+1},\bm\delta^{n+1},\bm\phi) \\
=& \underset{\bm\phi\in \mathcal{B}}{\arg\min} \left\{ 
\displaystyle\sum_{j=1}^{N}\left[ -\sum_{k=1}^{K} \phi_{jk} \ln f_{jk}(x_j^{n+1},\delta_{jk}^{n+1}) + \sum_{k=1}^K \phi_{jk} \ln \phi_{jk} \right]% 
\right\}.
\end{array}
\end{equation}
\STATE 5. If $\frac{||\bm{x}^{n+1}-\bm{x}^{n}||_2}{||\bm{x}^{n}||_2}<10^{-4}$, stop. Or let $n=n+1$, goto 2.
\end{algorithmic}
\end{algorithm}

We list the detailed algorithm in Algorithm \ref{alg2}. The only difference between Algorithm \ref{alg1} and Algorithm \ref{alg2} is the added Tikhonov regularization term in the sub-problem \eqref{x_sub_problem_reg} with respect to $\bm x$. With Tikhonov regularization the objective function is still quadratic, so we can apply CGLS method \cite{bjorck1996numerical} to solve \eqref{x_sub_problem_reg} efficiently. 

\subsection{Convergence results}
The model \eqref{model_ori_emsolver} could be seen as a special case of \eqref{model_regx} if we choose $\lambda_t = 0$. Thus, we only give the convergence results to Algorithm 2 for solving the minimization problem in \eqref{model_regx}.
\begin{proposition}
For the sequence $\{(\bm{x}^n,\bm{\delta}^n,\bm\phi^n)\}$ generated by 
Algorithm \ref{alg2}, every cluster point is a coordinatewise minimum
point of $F(\bm{x},\bm\delta,\bm\phi)$.

\end{proposition}

\begin{proof} We use Theorem 5.1 in \cite{Tseng2001} to give our convergence result. For self-contained, we list Theorem 5.1 in the appendix. For more details, please see \cite{Tseng2001}. Here we verify that our algorithm satisfies the assumptions of the theorem.

According to \eqref{model_regx},  
we set 
\begin{align*}
&F_0(\bm{x},\bm\delta,\bm\phi) = \sum_{j=1}^{N}\left[ -\sum_{k=1}^{K} \phi_{jk} \ln f_{jk}(x_j,\delta_{jk}) + \sum_{k=1}^K \phi_{jk} \ln \phi_{jk} \right],\notag\\
&F_1(\bm{x}) =\lambda_{n}||\bm A\bm x-\bm b||^2_{2} + \lambda_{t} \|\nabla \bm x\|^{2}_{2},\notag\\
&F_2(\bm\delta) = 
\left\{
\begin{array}{ll}
\lambda_{c} \sum_{k=1}^{K} R_{TV}(\bm\delta_k)& \text{ if } \sum_{k=1}^K \delta_{jk}=1, \delta_{jk}>0,\\
\infty& \text{ otherwise},
\end{array}
\right.\notag\\
&F_3(\bm\phi) = 
\left\{
\begin{array}{ll}
0 & \text{ if } \sum_{k=1}^K \delta_{jk}=1, \delta_{jk}>0,\\
\infty& \text{ otherwise}.
\end{array}
\right.\notag\\
\end{align*}

We have the following statements:

\begin{itemize}
\item The essentially cyclic rule (Definition \ref{appendix_ecr}) is satisfied, according to 
Algorithm \ref{alg2}.
\item Assumption B1 (in Definition \ref{def_b123c12}) is satisfied, due to that $F_0$ is continuous in its domain.
\item Assumption B2 (in Definition \ref{def_b123c12}) is satisfied, because the right hand of 
\eqref{x_sub_problem_reg} \eqref{delta_sub_problem_reg} \eqref{phi_sub_problem_reg} are strictly convex, by the following facts:
\begin{itemize}
\item 
\eqref{x_sub_problem_reg} is quadratic to $\bm{x}$ and the second term is strictly convex.
\item TV norm is convex in \eqref{delta_sub_problem_reg}.
\item The second term in \eqref{delta_sub_problem_reg} and the second term in \eqref{phi_sub_problem_reg} are separable, and each of the components is strictly convex.
\item The first term in \eqref{phi_sub_problem_reg} is linear.
\item The feasible sets of $F_1,F_2,F_3$ are convex.
\end{itemize}
\item Assumption B3 (in Definition \ref{def_b123c12}) is satisfied, because $F_0$ is continuous on its domain and $F_1,F_2,F_3$ are lsc.
\item Assumption C2 (in Definition \ref{def_b123c12}) is satisfied, because the domain of $F_0$ is $\mathbb{R}^N\times \mathcal{B}\times\mathcal{B}$.
\end{itemize}

By applying Theorem \ref{theorem_convergence} in Appendix, i.e., Theorem 5.1 in \cite{Tseng2001}, we know that for the sequence $\{(\bm{x}^n,\bm{\delta}^n,\bm\phi^n)\}$ generated by Algorithm \ref{alg2}, every cluster point is a coordinatewise minimum
point of $F$.

\end{proof}

\section{Numerical experiments}\label{section_exp}

In this section, we present some numerical experiment results to demonstrate the performance of our methods. We compare them with the one introduced in \cite{doi:10.1080/17415977.2015.1124428}, where a simplified model is solved instead of (\ref{model_SRS_ori}). All numerical tests are done on a linux server equipped with CPU 2.30Hz and MATLAB R2018a. In the tests, when we use CGLS method to solve the sub-problem with respect to $\bm x$ in \eqref{model_ori_emsolver} or \eqref{model_regx}, we set the maximum iteration number as 100 and the stopping rule as 
\[
\frac{\|\bm x^{m+1}-\bm x^{m}\|_{2}}{\|\bm x^{m}\|_{2}}\leq 10^{-4},
\]
where $m$ is the inner iteration index. 
When we solve the sub-problem on $\bm \delta$, we stop ADMM after 50 iterations or the condition 
\[
\frac{||\bm\delta^{m+1}-\bm\delta^{m}||_2}{||\bm\delta^m||_2}<10^{-4}
\]
is satisfied. In the split Bregman method for solving \eqref{subproblem_delta_delta}, the stopping rule is 
\[
\frac{||\bm\delta^{l+1}-\bm\delta^{l}||_2}{||\bm\delta^l||_2}<10^{-2},
\]
where $l$ is the iteration index inside split Bregman method. The parameter $\epsilon$ is set to 0.0001 for all experiments. 

\subsection{Experimental settings}
In our numerical experiments, the phantoms are generated from AIR Tools package \cite{hansen2012air} with the command \texttt{phantomgallery}, and the system matrix of CT scanner is obtained by calling \texttt{paralleltomo}. Without special mention, the size of phantoms is 64-by-64, the number of pixels in the detector is 91, and the projection angles are from $6^{\circ}$ to $180^{\circ}$ with the equal space $6^{\circ}$. Then, the underdetermined rate of our inverse problem, $A\bm x=\bm b$ is 0.667. In our tests, we assume that the measurements are corrupted by additive white Gaussian noise with mean 0 and standard deviation $\varepsilon\|A\bar{\bm x}\|_2$, where $\varepsilon$ gives the noise level and $\bar{\bm x}$ denotes the true attenuation coefficients. 

In order to illustrate the performance of the methods, we define the reconstruction error and segmentation error as 
\begin{equation}\label{error}
rec_{err} = \frac{||\bm{x}-\bar{\bm{x}}||_2}{||\bm{x}||_2}, \quad  seg_{err} = \frac{1}{N} \sum_{j\in\Omega} I(l_j-l_j^*),
\end{equation}
where $\bm{x}$ is the reconstruction result, $I(\cdot)$ is the Dirac function, $l_j$ is the label of pixel $j$ in the segmentation result, which is given by 
\begin{equation}\label{classification_l}
l_j = \underset{k}{\arg\max}\, \delta_{jk},
\end{equation}
and $l_j^*$ denotes the true label for pixel $j$. The regularization parameters in our method as well as in the method proposed in \cite{doi:10.1080/17415977.2015.1124428} are chosen as the ones which give the smallest $rec_{err}+seg_{err}$ value.

\subsection{Comparison on a piecewise constant phantom}\label{section_exp1}

We first compare our methods by solving (\ref{model_ori_emsolver}) and (\ref{model_regx}) with the one proposed in \cite{doi:10.1080/17415977.2015.1124428} on an 8-class piecewise constant phantom. The noise level $\varepsilon$ is set as 0.05, and the prior information on the mean and standard deviation for each class is $\mu_{k}=\frac{k-1}{7}$ and $\sigma_{k}=0.1$ for $k=1, \cdots, 8$. In our methods, we set the parameters as $\lambda_{n} = 0.2, \lambda_{c} = 1, \gamma_1 = 1, \gamma_2 = 2, \lambda_{t} = 1$. 

In Table \ref{table1}, we list the reconstruction error and segmentation error defined as in \eqref{error} together with CPU times in second for all three methods. All results shown in Table \ref{table1} are the average of running on 50 different noise realizations. 
It can be seen that the methods by solving (\ref{model_ori_emsolver}) and (\ref{model_regx}) can achieve almost the same segmentation results, and they are better than the one from \cite{doi:10.1080/17415977.2015.1124428}. 
Comparing reconstruction error, we can see that the method by solving (\ref{model_ori_emsolver}) gives the largest error, which is due to not enough regularization on the reconstruction. By adding Tikhonov regularization in the model (\ref{model_regx}), we are able to obtain comparable results as the simplified model in \cite{doi:10.1080/17415977.2015.1124428}. 
In addition, both our methods cost similar CPU time, which is less than the one in \cite{doi:10.1080/17415977.2015.1124428} with around a factor $\frac15$. 

\begin{table}[t]
\center
\caption{Comparison on a piecewise constant phantom}
\label{table1}
\begin{tabular}{cccc}
\hline\noalign{\smallskip}
  & $rec_{err}$ & $seg_{err}$ &  CPU Time (in second) \\
\noalign{\smallskip}\hline\noalign{\smallskip}
Method in \cite{doi:10.1080/17415977.2015.1124428}  &0.086& 0.033&469.7\\
\noalign{\smallskip}\hline\noalign{\smallskip}
Method by solving (\ref{model_ori_emsolver})  &0.106& 0.027& 93.2\\
Method by solving (\ref{model_regx})  &0.088& 0.026& 93.9\\
\noalign{\smallskip}\hline
\end{tabular}
\end{table}

\begin{figure}[t]
\begin{center}
\includegraphics{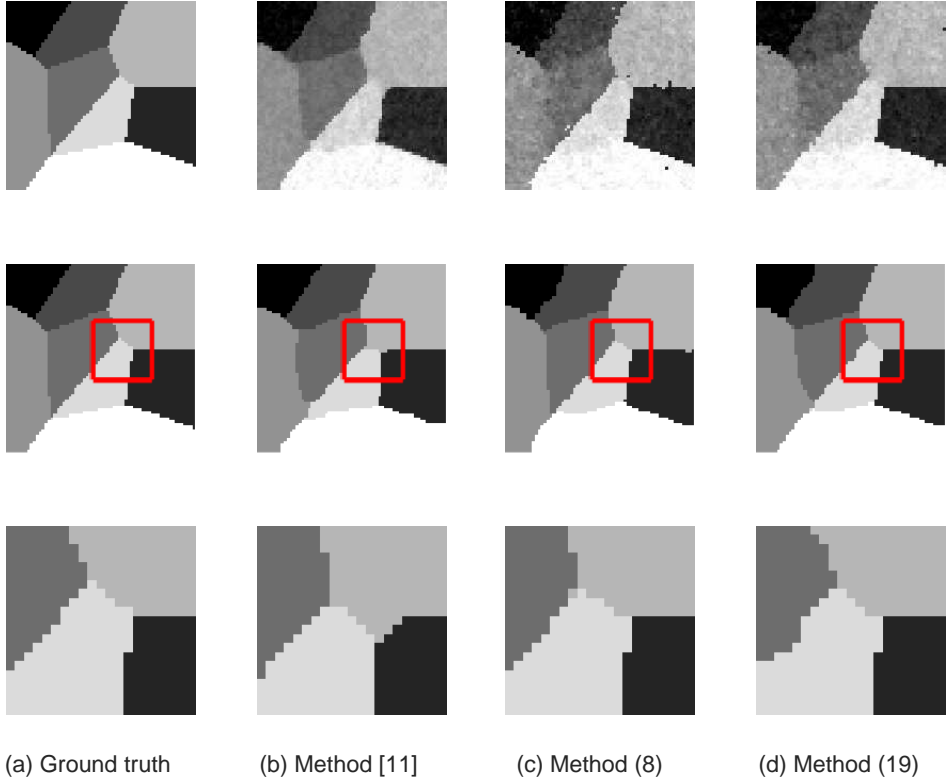}
\end{center}
\caption{Comparison of reconstruction and segmentation results with respect to the best case on a piecewise constant phantom. Row 1: reconstruction results; row 2: segmentation results; row 3: zoomed segmentation results in the red square.}
\label{8classbest}
\end{figure}

\begin{figure}[t]
\begin{center}
\includegraphics{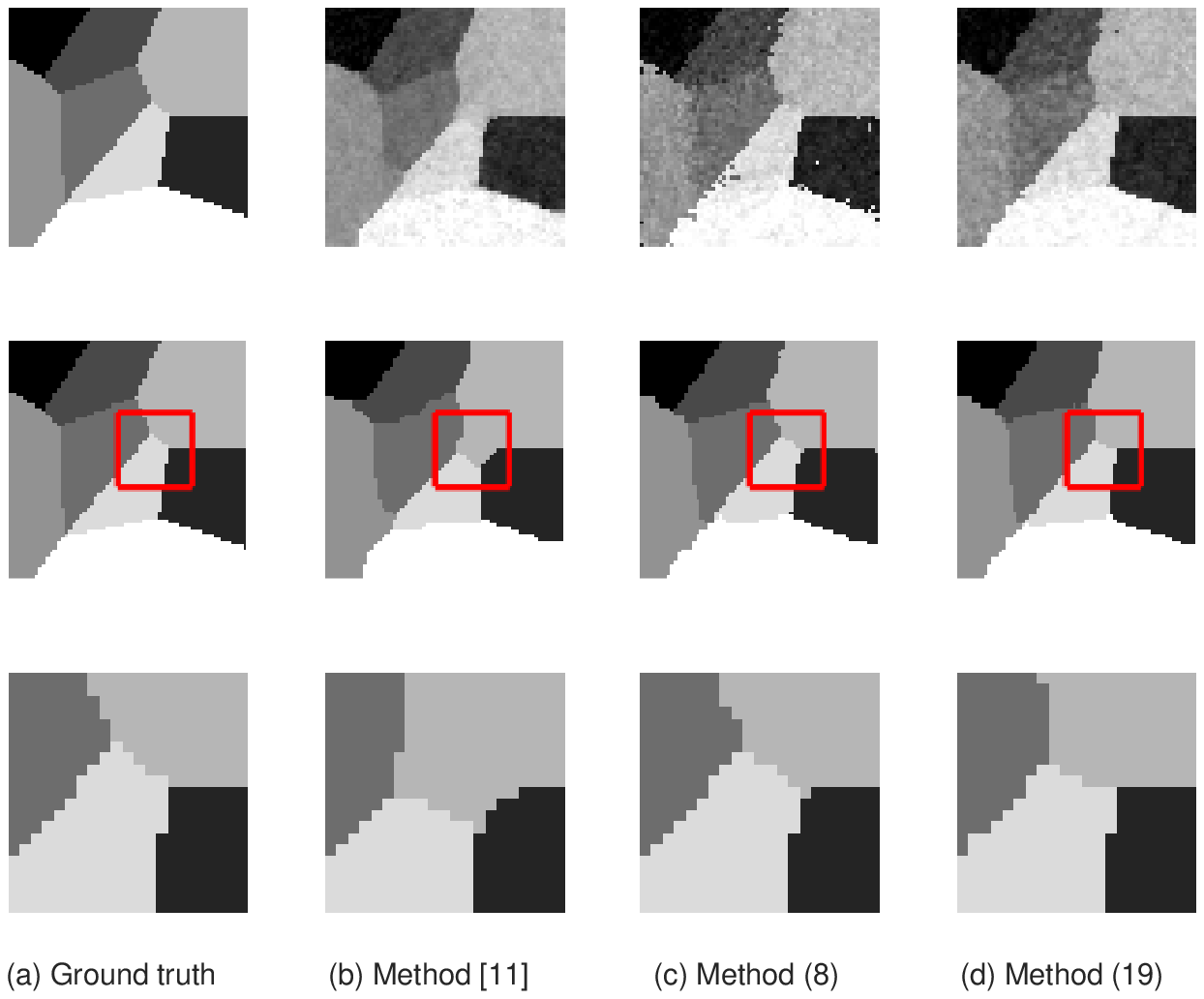}
\end{center}
\caption{Comparison of reconstruction and segmentation results with respect to the worst case on a piecewise constant phantom. Row 1: reconstruction results; row 2: segmentation results.}
\label{8classworst}
\end{figure}

In order to compare the results visually, in Figure \ref{8classbest} and \ref{8classworst} we show the reconstruction and segmentation results with respect to the best and worst case in these 50 tests, i.e., the results with the smallest and the largest $rec_{err}+seg_{err}$ values, respectively. It is clear that our methods provide more accurate segmentation results, which can be seen from the dark gray class in the middle red region, see the third row for zooming images. For the reconstruction results, the method in \cite{doi:10.1080/17415977.2015.1124428} gives much smoother results than both the proposed methods.

\begin{figure}[t]
\begin{center}
\includegraphics{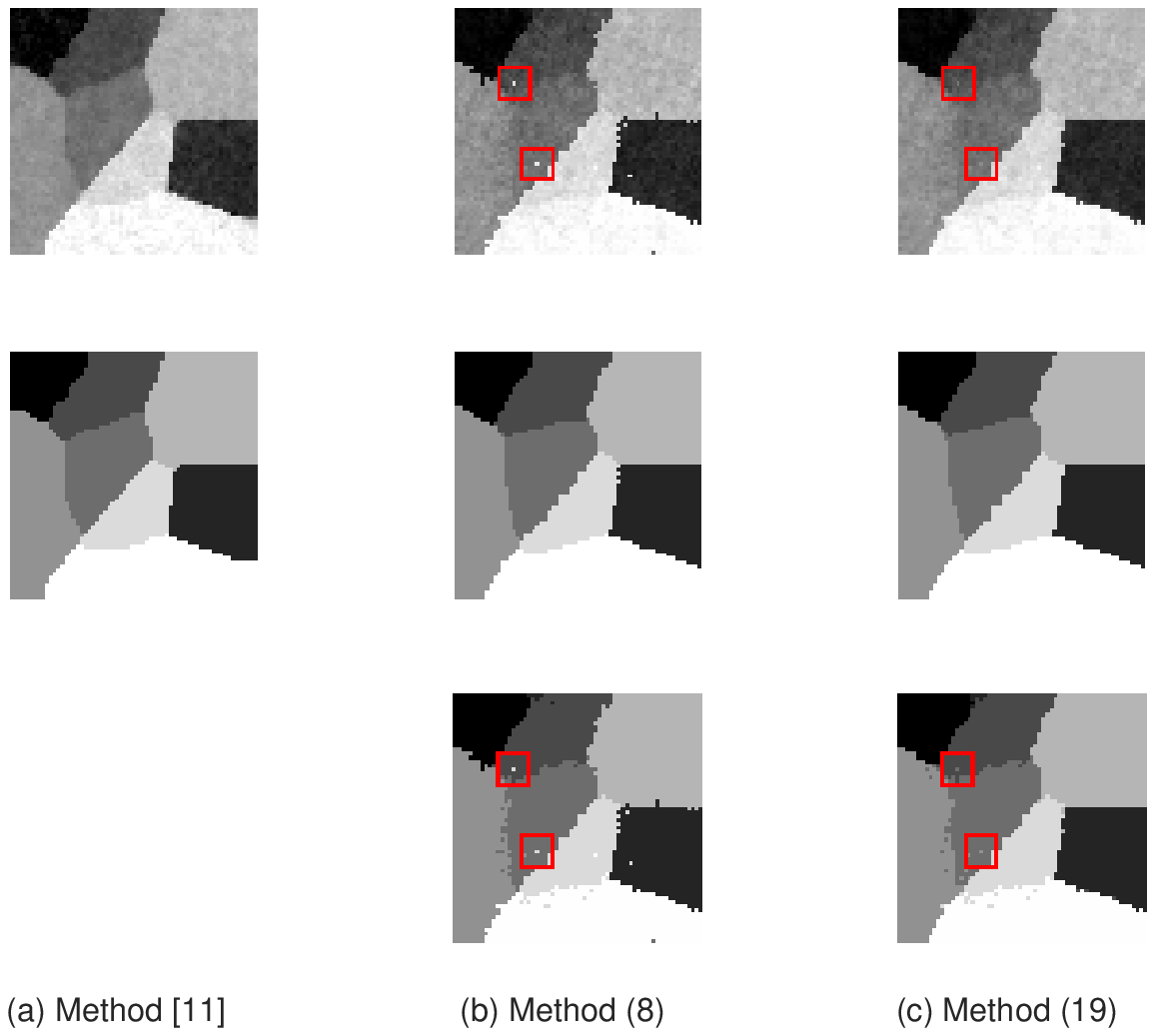}
\end{center}
\caption{Comparison of $\bm{x}, \bm\delta,\bm\phi$ in (\ref{model_ori_emsolver}), \eqref{model_regx} with $\bm{x}, \bm\delta$ in \cite{doi:10.1080/17415977.2015.1124428}. Row 1: the reconstructed $\bm{x}$; row 2: the segmentation from $\bm\delta$; row 3: the segmentation from $\bm\phi$.}
\label{8classcompareseg}
\end{figure}

In Figure \ref{8classcompareseg}, we use one example from the 50 tests to show the difference on $\bm{x}, \bm\delta,\bm\phi$ in (\ref{model_ori_emsolver}), \eqref{model_regx} comparing with $\bm{x}, \bm\delta$ in \cite{doi:10.1080/17415977.2015.1124428}. We treat $\bm\phi$ as a segmentation using \eqref{classification_l}, and show it in the third row of Figure \ref{8classcompareseg}. Note that in the method proposed in \cite{doi:10.1080/17415977.2015.1124428}, there is no variable $\bm\phi$.
As you can see, in method (\ref{model_ori_emsolver}) the isolated points in $\bm\phi$ marked by red squares lead to isolated points in the reconstructions. After adding regularization on $\bm{x}$ in (\ref{model_regx}), the number of isolated points is obviously reduced, and we can clearly see the smoothness in the reconstruction result from method \eqref{model_regx}. This indicates that the regularization term is helpful for improving the reconstruction. According to the reconstruction from \cite{doi:10.1080/17415977.2015.1124428}, although there is no regularization on $\bm x$, the reconstruction is still rather smooth without isolated points. It means that the simplification of the model in \cite{doi:10.1080/17415977.2015.1124428} potentially gives smoothing regularization on the reconstruction through the segmentation $\bm\delta$.   

\subsection{Comparison on a smooth phantom}
In this experiment, we compare the methods on a smooth phantom with 3 different classes, which is much more difficult to segment comparing with piecewise constant phantoms. The noise level is $\varepsilon = 0.01$, and the prior information on the mean and standard deviation for three classes are $\mu_{1} = 0.16$, $\mu_{2}=0.24$, $\mu_{3}=0.565$ and $\sigma_{k} = 0.05, k=1,2,3$. Other parameters are set as $\lambda_{n} = 123, \lambda_{c} = 0.55, \gamma_1 = 0.6, \gamma_2 = 0.6, \lambda_{t} = 35$. 

In Table \ref{table2}, we list the averages of the reconstruction errors, segmentation errors and CPU times on 50 experiment tests with different noise realizations. We can see that our methods by solving the models (\ref{model_ori_emsolver}) and (\ref{model_regx}) achieve similar segmentation results, and the method proposed in \cite{doi:10.1080/17415977.2015.1124428} provides the smallest segmentation error. Comparing the reconstruction results, it turns out that our method with the model (\ref{model_regx}) gives the best reconstruction results, followed by \cite{doi:10.1080/17415977.2015.1124428}. It is obvious that Tikhonov regularization in the model (\ref{model_regx}) plays an important role, which reduces the reconstruction error significantly. In addition, our methods cost only $\frac{1}{11}$ CPU time comparing with the method with simplified model. 

\begin{table}[t]
\center
\caption{Comparison on a smooth phantom}
\label{table2}
\begin{tabular}{cccc}
\hline\noalign{\smallskip}
  & $rec_{err}$ & $seg_{err}$ &  CPU Time (in second) \\
\noalign{\smallskip}\hline\noalign{\smallskip}
Method in \cite{doi:10.1080/17415977.2015.1124428} & 0.203&0.166& 374.9\\
\noalign{\smallskip}\hline\noalign{\smallskip}
Method by solving (\ref{model_ori_emsolver}) &0.215&0.177& 28.8 \\
Method by solving (\ref{model_regx}) &0.195&0.172 & 33.7\\
\noalign{\smallskip}\hline
\end{tabular}
\end{table}

\begin{figure}[t]
\begin{center}
\includegraphics{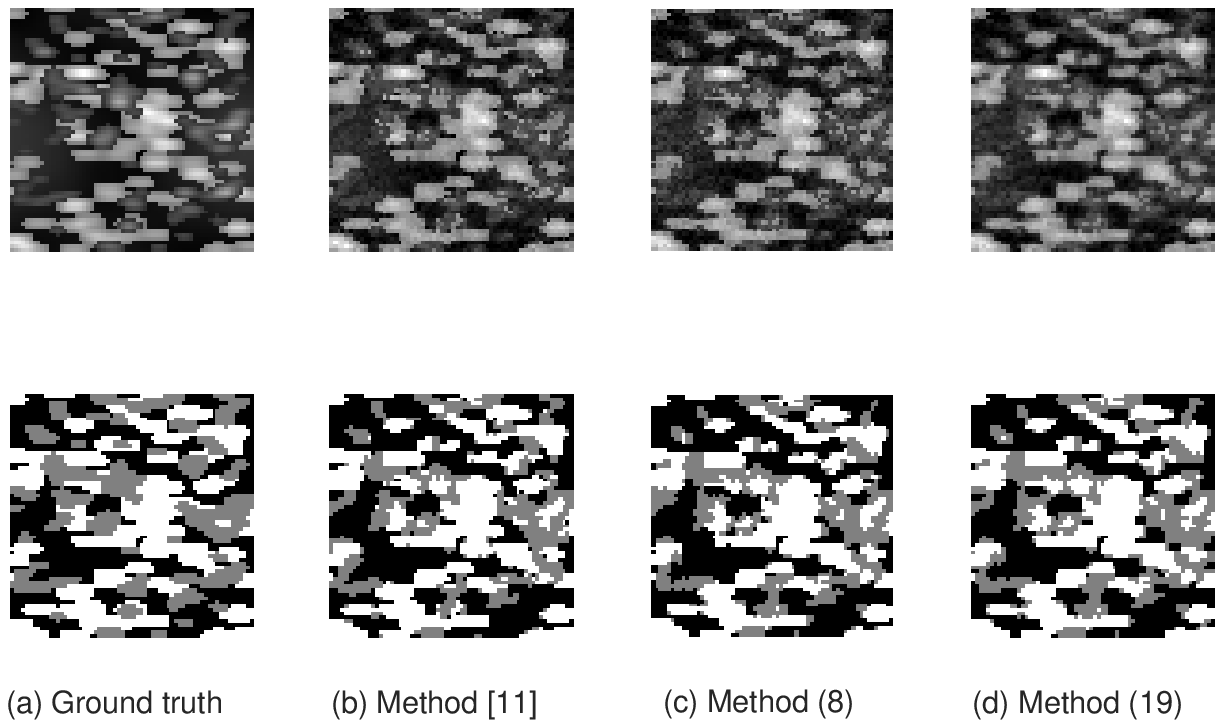}
\end{center}
\caption{Comparison of reconstruction and segmentation results with respect to the best case on a smooth phantom. Row 1: reconstruction results; row 2: segmentation results.}
\label{3classbest}
\end{figure}

\begin{figure}[t]
\begin{center}
\includegraphics{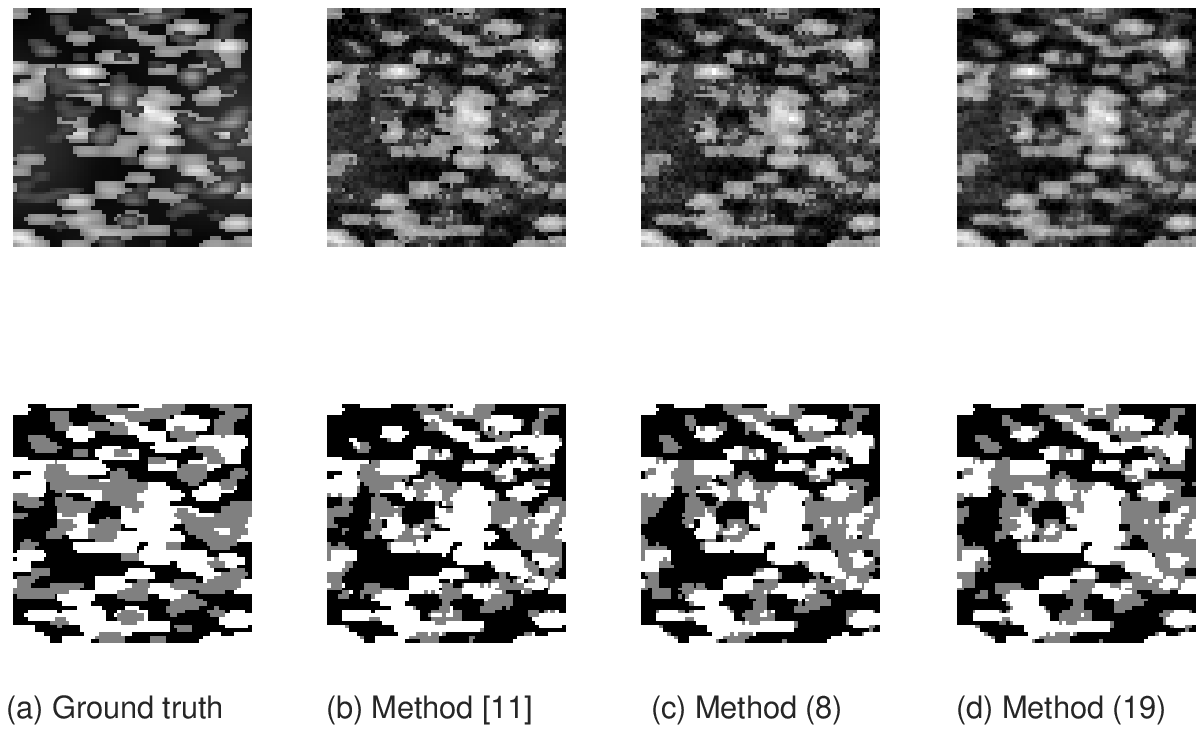}
\end{center}
\caption{Comparison of reconstruction and segmentation results with respect to the worst case on a smooth phantom. Row 1: reconstruction results; row 2: segmentation results.}
\label{3classworst}
\end{figure}

In Figure \ref{3classbest} and \ref{3classworst} we show the reconstruction and segmentation results from three methods under the best case and the worst case. The three methods generate comparable results. As you can see, there are no big differences between the three methods.

\subsection{Comparison of CPU times on different resolutions}\label{sec_pdis}

In this section, we compare our methods with the method in \cite{doi:10.1080/17415977.2015.1124428} with respect to CPU times. 
To do so, we generate the piecewise constant phantom with different resolutions, then adjust the number of the projection angles such that the underdetermined rate is still kept as 0.667. The test results are the averages on 3 different noise realizations.

\begin{table}[t]
\center
\caption{Comparison of CPU time (in second) on a piecewise phantom}
\label{table3}
\resizebox{\textwidth}{!}{
\begin{tabular}{ccccc}
\hline\noalign{\smallskip}
phantom resolution & $64\times 64$ & $128\times 128$ & $256\times 256$ &$512\times 512$ \\
\noalign{\smallskip}\hline\noalign{\smallskip}
Projection angle & $6^{\circ}:6^{\circ}:180^{\circ}$ & $3^{\circ}:3^{\circ}:180^{\circ}$&$1.5^{\circ}:1.5^{\circ}:180^{\circ}$&$0.75^{\circ}:0.75^{\circ}:180^{\circ}$ \\
\noalign{\smallskip}\hline\noalign{\smallskip}
Method in \cite{doi:10.1080/17415977.2015.1124428}  & 469.7 & 1981.4&9447.3& --\\
\noalign{\smallskip}\hline\noalign{\smallskip}
Method with (\ref{model_ori_emsolver}) & 93.2 &161.2&783.5& 3224.2\\
Method with (\ref{model_regx}) & 93.9 &138.0&560.4&2928.7\\
\noalign{\smallskip}\hline
\end{tabular}
}
\end{table}

In Table \ref{table3}, we list the CPU times in second for all three methods. Note that in the case of 512-by-512 phantom, we could not apply the method in \cite{doi:10.1080/17415977.2015.1124428} due to heavy computational cost and limited memory. It is obvious that both our methods cost much less CPU times than the one in \cite{doi:10.1080/17415977.2015.1124428}. When the resolution increases, the method with Tikhonov regularization, i.e., the one solving the model \eqref{model_regx}, utilizes less and less computing time compared to the one in \cite{doi:10.1080/17415977.2015.1124428}, with a factor of $\frac{1}{5},\frac{1}{14},\frac{1}{16}$, and less time compared to method (\ref{model_ori_emsolver}). 
The latter one is because, the sub-problem on $\bm x$ is better conditioned and less iterations are needed in order to reach stopping rule. 

All the above comparison indicate that the proposed algorithm is faster than the method in \cite{doi:10.1080/17415977.2015.1124428}. There might be three reasons: a, the objective function of the Frank-Wolf algorithm is singularity due to the TV norm; b, the dimension of $\bm{x}$ is large; c, although there are more variables to solve in the proposed method, the proposed method is separable and many sub-problems have close-form solutions. 

\section{Conclusion}\label{section_con}
In this paper, we propose two new methods for simultaneous reconstruction and segmentation in X-ray CT application.
By using the commutativity of log-sum operations, the original model proposed in \cite{doi:10.1080/17415977.2015.1124428} could be solved more efficiently. Although one more variable is introduced, the energy function becomes separable and each sub-problem is convex and easy-to-solve. We also show some convergence results, and numerically discover the role of the simplification steps in \cite{doi:10.1080/17415977.2015.1124428}. Numerical results show that the proposed methods provide comparable reconstruction and segmentation results with much less CPU time.   

\section{Acknowledgments} 
We thank Hans Martin Kjer from Technical University of Denmark for providing us the codes for the method introduced in \cite{doi:10.1080/17415977.2015.1124428}. Y. Dong acknowledges the support of the National Natural Science Foundation of China under Grant 11701388 and the Villum Foundation under Grant 25893. C. Wu was supported by National Natural Science Foundation of China under Grants 11871035 and 11531013; and Recruitment Program of Global Young Expert.

\bibliographystyle{tfnlm}
\bibliography{bibtex}

\section*{Appendix}
All the following definitions and theorem come from \cite{Tseng2001}.

Let $f$ have the following form:
\begin{equation}
\min f(\bm{x}_1,\cdots,\bm{x}_n) = f_0(\bm{x}_1,\cdots,\bm{x}_n) + \sum_{k=1}^N f_k(\bm{x}_k).
\end{equation}
where $f_0: \mathbb{R}^{n_1 + \cdots + n_N}\rightarrow \mathbb{R} \cup \{\infty\}$, $f_k: \mathbb{R}^{n_k}\rightarrow \mathbb{R} \cup \{\infty\}$. $f$ is proper. 
\begin{definition}
Block coordinate descent (BCD) method:
\begin{itemize}
\item Initialization. Choose any $\bm{x}^0 = (\bm{x}_1^0,\cdots,\bm{x}_N^0)\in dom f$.
\item Iteration $r+1$ with $r\geq 0$. Given $\bm{x}^r = (\bm{x}_1^r,\cdots,\bm{x}_N^0)\in dom f$, choose an index $s\in\{1, . . . , N\}$ and compute a new iterate
$$\bm{x}^{r+1} = (\bm{x}_1^{r+1},\cdots,\bm{x}_N^{r+1})\in dom f,$$
satisfying 
$$\bm{x}_s^{r+1} \in \arg\underset{\bm{x}_s}{ \min} f(\bm{x}_1^r,\cdots,\bm{x}_{s-1}^r, \bm{x}_s , \bm{x}_{s+1}^r,\cdots,\bm{x}_N^r),$$
$$\bm{x}_j^{r+1} = \bm{x}_j^r, \forall j\neq s.$$
\end{itemize}
\end{definition}

\begin{definition}\label{appendix_ecr}
[Essentially cyclic rule] There exists a constant $T\geq N$ such that every
index $s\in\{1, . . . , N\}$ is chosen at least once between the $r$th iteration and the
($r+T-1$)th iteration, for all $r$.
\end{definition}

\begin{definition}\label{def_b123c12}
More assumptions about $f,f_0,\cdots, f_{N}$.
\begin{itemize}
\item (B1) $f_0$ is continuous on $dom f_0$.
\item (B2) For each $k\in\{1, . . . , N\}$ and $(\bm{x}_j)_{j\neq k}$, the function $\bm{x}_k\rightarrow f(\bm{x}_1,\cdots,\bm{x}_N)$ is quasiconvex and hemivariate.
\item (B3) $f_0,f_1,\cdots,f_N$ are lsc.
\item (C1) $dom f_0$ is open and $f_0$ tends to $\infty$ at every boundary point of $dom f_0$.
\item(C2) $dom f_0 = Y_1\times \cdots \times Y_N$, for some $Y_k\subseteq \mathbb{R}^{n_k}, k=1,\cdots,N$.
\end{itemize}
\begin{itemize}
\item $h$ is quasiconvex if
$$h(\bm{x}+\lambda \bm{d})\leq max\{h(\bm{x}),h(\bm{x}+\bm{d})\}, \forall \bm{x},\bm{d}, \lambda\in[0,1].$$
\item $h$ is hemivariate if $h$ is not constant on any line
segment belonging to $dom h$.
\item $\bm{z}$ is a coordinatewise minimum point of $f$, if $\bm{z}\in dom f$ and 
$$f(\bm{z}+(0,\cdots,d_k,\cdots,0))\geq f(\bm{z}), \forall d_k \in \mathbb{R}^{n_k}, \forall k.$$
\item if $h$ is strictly convex, then $h$ is quasiconvex and hemivariate.
\end{itemize}
\end{definition}

\begin{theorem}[Theorem 5.1 in \cite{Tseng2001}]\label{theorem_convergence}
Suppose that $f,f_0,\cdots,f_N$ satisfy Assumptions B1, B2, B3 and that $f_0$ satisfies either Assumption C1 or C2. 
Also, assume that the
sequence $\{\bm{x}^r = (\bm{x}_1^r,\cdots,\bm{x}_N^r)\}_{r=0,1,\cdots}$ generated by the BCD method using
the essentially cyclic rule is defined. Then, either $\{f(\bm{x}^r)\}\rightarrow -\infty$, or else every
cluster point $\bm{z} = (\bm{z}_1,\cdots,\bm{z}_N)$ is a coordinatewise minimum point of $f$.
\end{theorem}

\end{document}